\pdfoutput=1

\documentclass[11pt]{article}

\usepackage[]{emnlp2021}

\usepackage{times}
\usepackage{latexsym}
\usepackage{graphicx} 
\usepackage[T1]{fontenc}
\usepackage{multirow}
\usepackage{booktabs}
\usepackage{amsmath}

\usepackage[utf8]{inputenc}

\usepackage{microtype}
\usepackage{url}

\usepackage[export]{adjustbox} 

\newcommand{\et}{\emph{et al. }}

\newcommand{\tabincell}[2]{\begin{tabular}{@{}#1@{}}#2\end{tabular}}
\title{Wav-BERT: Cooperative Acoustic and Linguistic Representation Learning for Low-Resource Speech Recognition }

\author{Guolin Zheng\textsuperscript{\rm 1}\thanks{* Both authors contributed equally to this work.} , Yubei Xiao\textsuperscript{\rm 1}\footnotemark[1] , Ke Gong\textsuperscript{\rm 2}, Pan Zhou\textsuperscript{\rm 3}, Xiaodan Liang\textsuperscript{\rm 1}, Liang Lin\textsuperscript{\rm 1,2}\thanks{Liang Lin is the corresponding author of this work.}  \\
\textsuperscript{\rm 1}Sun Yat-sen University, China \\
\textsuperscript{\rm 2}Dark Matter AI Research, \textsuperscript{\rm 3}Sea AI Lab \\
\texttt{\{zhengglin,xiaoyb5\}@mail2.sysu.edu.cn} \\
\texttt{kegong936@gmail.com}, \texttt{zhoupan@sea.com}, \\
\texttt{xdliang328@gmail.com}, \texttt{linliang@ieee.org}}

\begin{document}
\maketitle
\begin{abstract}

Unifying acoustic and linguistic representation learning has become increasingly crucial to transfer the knowledge learned on the abundance of high-resource language data for low-resource speech recognition.  Existing approaches simply cascade pre-trained acoustic and language models to learn the transfer from speech to text. However, how to solve the representation discrepancy of speech and text is unexplored, which hinders the utilization of acoustic and linguistic information. Moreover, previous works simply replace the embedding layer of the pre-trained language model with the acoustic features,  which may cause the catastrophic forgetting problem. In this work, we introduce Wav-BERT, a cooperative acoustic and linguistic representation learning method to fuse and utilize the contextual information of speech and text.  Specifically, we unify a pre-trained acoustic model (wav2vec 2.0) and a language model (BERT) into an end-to-end trainable framework. A Representation Aggregation Module is designed to aggregate acoustic and linguistic representation, and an Embedding Attention Module is introduced to incorporate acoustic information into BERT, which can effectively facilitate the cooperation of two pre-trained models and thus boost the representation learning. Extensive experiments show that our Wav-BERT significantly outperforms the existing approaches and achieves state-of-the-art performance on low-resource speech recognition.

\end{abstract}

\section{Introduction}
Recently, Automatic Speech Recognition (ASR) has achieved  remarkable success, which can be attributed to two complementary aspects: 1) designing more effective and larger deep neural networks for ASR, and 2) training on a large amount of   data~\cite{chan2016listen,watanabe2017hybrid,amodei2016deep}.   However, in practice, unlike the commonly used languages (e.g. English and Chinese) with sufficient training data, many other languages (e.g. Swahili, Tamil) have only low-resource data due to  the scarcity of audios and the huge labor resources consumed in transcription. In this way,  the aforementioned  data-driven mechanism is impractical for low-resource languages and thus suffers from unsatisfactory performance. 

To resolve this learning difficulty in the low-resource domain, many efforts have been devoted to leveraging unlabeled data. One mainstream research paradigm is unsupervised pre-training, or representation learning, which has achieved great  success in natural language processing~\cite{devlin2018bert,peters2018deep} and received increasing attention in speech recognition~\cite{oord2018representation,schneider2019wav2vec}. 
As a representation in this line, wav2vec~\cite{schneider2019wav2vec} and wav2vec 2.0~\cite{baevski2020wav2vec}  apply  unsupervised contrastive pre-training and show promising results. To utilize linguistic information, some works~\cite{chiu2021innovative,shin2019effective} also aim to  build  language models to rescore the $N$-best hypotheses generated by acoustic models. The most recent approach~\cite{yi2021efciently} even cascaded the pre-trained wav2vec 2.0 and BERT into a single model for low-resource ASR. 

However, there leave two critical challenges on how to integrate the acoustic model and language model to utilize the contextual information of speech and text. 1) Representation discrepancy: the acoustic model focuses more on local dependencies of the speech sequence, while the language model aims at capturing long-term semantic information of texts. It is desired to explore an effective model to fuse and leverage the two kinds of representation. 2) Embedding inconsistency: The language model applies a token embedding layer during pre-training but previous methods~\cite{yi2021efciently} simply replace the embedding layer with the features generated by the acoustic model, which may result in the catastrophic forgetting problem~\cite{goodfellow2013empirical}.

To tackle the above challenges, in this work, we make the first attempt to successfully integrate the well-trained acoustic model and language model for low-resource speech recognition.  Towards this end, we introduce a new framework that incorporates the two kinds of pre-trained models for cooperative acoustic and linguistic representation learning by exploiting complementary contextual information of both speech and text.    

First, to solve representation discrepancy, unlike the previous works~\cite{yi2021efciently,yu2021non} that simply connect the acoustic model and the language model by treating them as an encoder and a decoder, we consider them as two encoders that provide two different representations. Specifically, we propose a Representation Aggregation Module, a plug-in component to better exploit and fuse the acoustic and linguistic information. We design and evaluate several representation aggregation mechanisms, including Gated Acoustic-Guided Attention, Gated Linguistic-Guided Attention, and Gated Cross-Modal Attention. The experimental results show the proposed Gated Cross-Modal Attention is the most effective method for representation aggregation.


Second, to fill the gap of embedding inconsistency, we introduce an Embedding Attention Module to incorporate the acoustic features into BERT by a gated attention process, which not only preserves the capability of BERT but also takes advantage of acoustic information. Moreover, as BERT requires audio transcripts as input to create word embedding, it may be easy to overfit when using ground truth transcripts. On the other hand, it is also hard to converge when using transcripts predicted by the acoustic model. To facilitate the cooperation of the two encoders, we propose a sampling strategy with decay to randomly select the ground truth and generated transcripts for smooth training.

We adopt pre-trained wav2vec 2.0~\cite{baevski2020wav2vec} and BERT~\cite{devlin2018bert} as the encoders to provide acoustic and linguistic representations respectively for their flexible pre-training then fine-tuning paradigm as well as excellent local contextual modeling ability. Accordingly, we denominate our method as Wav-BERT.

We evaluate  our method on several datasets with diverse languages from the public IARPA BABEL dataset~\cite{gales2014speech} and AISHELL-1 corpus~\cite{bu2017aishell}. The experimental results demonstrate that our Wav-BERT significantly outperforms the existing approaches on low-resource ASR. Furthermore, our  exhaustive ablation studies demonstrate  the effectiveness of the proposed   mechanisms for cooperative acoustic and linguistic representations learning. We hope this work will be useful for the community on the way to explore different pre-trained models for low-resource ASR.

\section{Related Work}
\subsection{Low resource speech recognition}
To tackle the low-resource ASR task, transfer learning ASR~\cite{Kunze2017TransferLF} and multilingual transfer learning ASR~\cite{Dalmia2018SequenceBasedML,Watanabe2017LanguageIE,Toshniwal2018MultilingualSR} are explored via using different source languages to improve the performance of low-resource languages. Meta-learning approaches~\cite{Finn2017ModelAgnosticMF,Nichol2018OnFM} are also adopted for low-resource ASR~\cite{Hsu2019MetaLF,xiao2020adversarial} to obtain fast adaptation ability to new tasks with only a few data through meta-learning a model initialization from training tasks. In addition, recent works utilize unsupervised pre-training~\cite{Schneider2019wav2vecUP,Chung2019GenerativePF} and semi-supervised learning~\cite{Kahn2019SelfTrainingFE,Li2019SemisupervisedTF} to exploit a large amount of unlabeled data to learn general representations for low-resource adaptation. 
Among them, Wav2vec 2.0~\cite{baevski2020wav2vec} achieved excellent results through self-supervised learning, which learns powerful and contextual acoustic representations of a large speech audio corpus by solving contrastive tasks that require identifying the true quantized latent speech representations for masked time steps.
Then it shows strong feasibility of ultra-low resource speech recognition with even only 10 minutes of labeled data.

\begin{figure*}

    \centering
    \includegraphics[width=1.0\linewidth, clip]{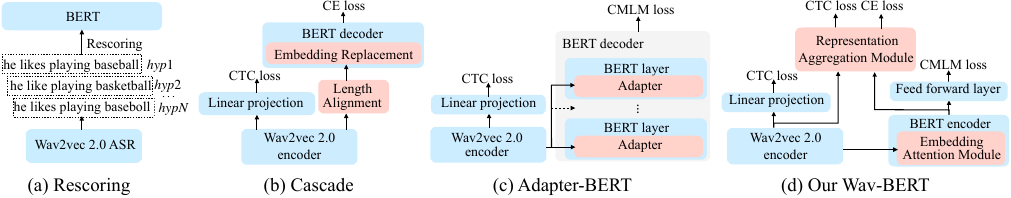}
    \caption{Comparison of the architectures of different approaches to fuse BERT into the ASR model. (a) Rescoring methods use BERT to rescore $N$-best hypotheses generated by wav2vec 2.0 ASR~\cite{shin2019effective}. (b) Cascade methods directly cascade the BERT decoder on the top of the wav2vec 2.0 encoder through Length Alignment module~\cite{yi2021efciently}. (c) Adapter-BERT inserts adapter modules in each BERT layer~\cite{NEURIPS2020_7a6a74cb}.
    (d) Our Wav-BERT introduces a Representation Aggregation Module for aggregate acoustic and linguistic representation and an Embedding Attention Module to incorporate acoustic information into text embedding.}
    \label{fig:model_fusion}

\end{figure*}

\subsection{Speech recognition with BERT}
To use the linguistic information from BERT~\cite{devlin2018bert} for improving ASR performance, some works~\cite{chiu2021innovative,shin2019effective,wang2019bert} use BERT to re-rank the N-best hypotheses generated by the ASR model.  Besides, knowledge distillation~\cite{futami2020distilling} is explored to use BERT as a teacher model to guide ASR model training. Moreover, some recent works~\cite{yi2021efciently,yu2021non,winata2020adapt} further combine BERT with the ASR model into a unified model and train the model in an end-to-end way. But Yi~\et  and Yu~\et  both simply connect BERT and the ASR model in series without considering the contextual information of speech and text~\cite{yi2021efciently,yu2021non}. Winata~\et~\cite{winata2020adapt} modified mBERT model into an auto-regressive decoder and insert a cross-attention layer in each mBERT layer, but the deep bidirectional information of pre-trained BERT cannot be fully utilized in the auto-regressive mode.

\section{Preliminaries}
Here we briefly introduce the architectures of acoustic and linguistic encoders in our framework.

\noindent\textbf{Wav2vec 2.0.}
We adopt  wav2vec 2.0~\cite{baevski2020wav2vec} as our acoustic encoder because of its effectiveness and efficiency. It has two stages: (i) contrastive pre-training to learn representations of speech  and  (ii) fine-tuning to adapt the learned representations on labeled data with connectionist temporal classification(CTC) loss~\cite{Graves2006ConnectionistTC} for downstream speech recognition tasks. In this work, we aim to utilize the public pre-trained model and mainly focus on the fine-tuning stage. 
The architecture of wav2vec 2.0 contains a feature encoder, a context network with a transformer and a quantization module. During fine-tuning, the quantization module is removed and a randomly initialized linear projection layer is attached on top of the context network. 

\noindent\textbf{BERT.}
  BERT~\cite{devlin2018bert} is employed as our linguistic encoder since it is one of the most popular text pre-training approaches and has shown remarkable performance in many downstream natural language processing tasks. It also consists of two steps: (i) self-supervised pre-training to learn deep bidirectional linguistic representations from a large text corpus  and (ii) fine-tuning to adapt to downstream tasks using labeled data.  BERT consists of an embedding table, a multi-layer bidirectional Transformer encoder, and an additional output layer for fine-tuning. 

\section{Wav-BERT}
\begin{figure*}
    \centering
    \includegraphics[width=1.0\linewidth,clip]{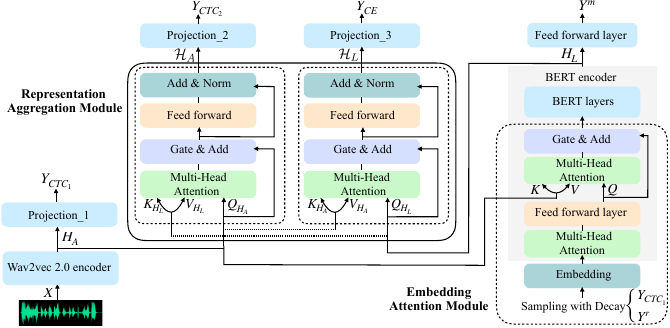}
    \caption{Our Wav-BERT framework, which is composed of two main parts: 1) Representation Aggregation Module that combines a Gated Acoustic-Guided Attention (Left) and a Gated Linguistic-Guided Attention (Right) to construct a Gated Cross-Modal Attention. 2) Embedding Attention Module that includes a Gated Attention and a "Sampling with Decay" mechanism. }
    \label{fig:overview}
\end{figure*}
\label{sec:out_method}
\subsection{Motivation}
\label{sec:model_fusion}

To transfer the knowledge learned on the abundance of high-resource language data for low-resource speech recognition, many efforts have been devoted to unifying acoustic and linguistic representation learning. We first categorize previous methods and then introduce our solution.

As shown in Figure~\ref{fig:model_fusion} (a), one simplest way to fuse BERT into an acoustic model in speech recognition is rescoring~\cite{chiu2021innovative,shin2019effective}. It uses BERT as a language model to calculate the pseudo-log-likelihood scores of text sentences for reranking the $N$-best hypotheses generated by the acoustic model. However, this process is time-consuming as it needs to iteratively mask each word in the sentence for inference and then sum up the scores of all masked words. It also requires tuning many hyper-parameters by repetitive experiments, e.g.  beam size, balanced weights of the language and acoustic models. 



Recently, some works~\cite{yi2021efciently,yu2021non} directly cascade  the decoder BERT  on the top of the acoustic encoder, as illustrated by  Figure~\ref{fig:model_fusion} (b). However, such a simple cascade often cannot well fuse the contextual information of speech and text.


Inspired by AB-Net~\cite{NEURIPS2020_7a6a74cb}, we design Adapter-BERT that inserts cross-attention adapters in each BERT layer with the Mask-Predict algorithm~\cite{ghazvininejad2019mask} to fully utilize the bidirectional information of the input sequence, as shown in Figure~\ref{fig:model_fusion} (c). Nevertheless, the adapters in each layer of BERT will affect the pre-trained parameters of BERT, causing catastrophic forgetting. Moreover, the Mask-Predict decoding suffers from low inference speed.

To solve the representation discrepancy and embedding inconsistency between speech and text, in this work, we introduce Wav-BERT, a cooperative acoustic and linguistic learning framework that fuses and leverages the contextual information of speech and text from the representation level to the embedding level,  as shown in Figure~\ref{fig:model_fusion} (d). 
We first present an independent Representation Aggregation Fusion Module for acoustic and linguistic representation aggregation, without inserting it in any pre-trained model to avoid destroying the parameters of pre-trained models.
Then, an Embedding Attention Module is introduced to better combine acoustic and linguistic embedding instead of simply replacement. 

\subsection{Our Wav-BERT}


The architecture of our Wav-BERT is illustrated in Figure~\ref{fig:overview}. 
Specifically, wav2vec 2.0 encoder takes raw waveform $X$ as input and outputs acoustic representation $H_A$, which is then fed into a linear projection layer with CTC loss~\cite{Graves2006ConnectionistTC} ($L_{ctc_1}$) and the Representation Aggregation Module respectively. 
For the input of BERT encoder, we employ ``Sampling with Decay" mechanism to sample from the masked ground truth $Y^r$ or wav2vec 2.0 CTC output $Y_{CTC_1}$ with probability $p$ and $1-p$, so as to narrow the gap between training and inference. Next, word embedding $E$ and acoustic embedding $H_A$ are fed into the Gate Attention to model the conditional information from the wav2vec 2.0 encoder side. Through the subsequent BERT transformer layers, we get the linguistic representation $H_L$. Finally, the Representation Aggregation Module takes linguistic representation $H_L$ as well as acoustic representation $H_A$ as input, generating the CTC output $Y_{CTC_2}$ and cross-entropy (CE) output $Y_{CE}$, supervised by the CTC ($L_{ctc_2}$) and CE ($L_{ce}$) criterion respectively. Simultaneously, the conditional masked language model (CMLM) objective ($L_{cmlm}$)~\cite{NEURIPS2020_7a6a74cb} is also attached on BERT encoder followed by a feed-forward layer to supervise the BERT output $Y^m$. Overall, the objective of our framework is defined as:

\begin{small}
\begin{equation}
  L= \mu_{1} \cdot L_{ctc_1} + \mu_{2} \cdot L_{cmlm} + \mu_{3} \cdot L_{ctc_2} + \mu_{4} \cdot L_{ce},
\end{equation}
\end{small}
where $\mu_{1}$, $\mu_{2}$, $\mu_{3}$ and $\mu_{4}$ are the corresponding loss weights.

\subsubsection{Representation Aggregation Module}
\label{sec:rep_agg}

To solve representation discrepancy, we first design several representation aggregation mechanisms, such as Gated Acoustic-Guided Attention, Gated Linguistic-Guided Attention. In our Representation Aggregation Module, we combine a Gated Acoustic-Guided Attention (Left) and a Gated Linguistic-Guided Attention (Right) to construct a Gated Cross-Modal Attention for better exploiting and aggregating the acoustic and linguistic representations.  

Specifically, Gated Cross-Modal Attention Module takes acoustic representation $H_A$ generated by wav2vec 2.0 as well as linguistic representation $H_L$ generated by BERT as input and feeds them as the query, key, and value vector respectively to a multi-head attention, which can be formulated as:
\begin{align}
   C_{A} &= \mbox{ATT}(Q_{H_A}, K_{H_L}, V_{H_L}), \\
   C_{L} &= \mbox{ATT}(Q_{H_L}, K_{H_A}, V_{H_A}),
\end{align}
where $Q_{H_A}$ means passing $H_A$ as query vector, $K_{H_L}$ as well as $V_{H_L}$ means passing $H_L$ as key and value vector respectively. $C_{A}$ is the acoustic guided context feature generated by attention which tend to focus on the values in the linguistic representation $H_L$ related to acoustic representation $H_A$.  Vice versa, $C_{L}$ is the linguistic guided context feature to focus on the values in the $H_A$ related to $H_L$.

Next, the context feature $C_{A}$ and acoustic representation $H_A$ are fed into a gated weighting layer to automatically capture the  most important information between context and acoustic representation, and generating acoustic-guided linguistic representation $H_{AGL}$, which can be   formulated as:
\begin{align}
    & \Phi_A = \mbox{sigmoid}(W_{1}[C_A;H_A] + B_{1}), \\
    & H_{AGL} = H_A + \Phi_{A} C_{A},
\end{align}
where  $W_{1}$ as well as  $B_{1}$ are model parameters and $\Phi_A $ is the gated weight.

Similarly, the context feature $C_{L}$ and linguistic representation $H_L$ are fed into another gated weighting layer to weigh the expected importance $\Phi_L$ and generate  linguistic-guided acoustic representation $H_{LGA}$, which can be   formulated as:
\begin{align}Tt
    & \Phi_L = \mbox{sigmoid}(W_{2}[C_L;H_L] + B_{2}), \\
    & H_{LGA} = H_L + \Phi_{L} C_{L},
\end{align}
where  $W_{2}$ as well as $B_{2}$ are model parameters and $\Phi_L $ is the gated weight.

We then feed  $H_{AGL}$ and $H_{LGA}$  to  a feed-forward layer followed by residual connection respectively and get aggregation representation  $\mathcal{H}_A$ as well as  $\mathcal{H}_L$. 
Finally, two linear projection layers are attached on the top of Representation Aggregation Module to get the $Y_{CTC_2}$ and $Y_{CE}$.
As the sequence length of $Y_{CTC_2}$ is determined by acoustic representation $H_A$, we use CTC criterion to align the acoustic frames of $Y_{CTC_2}$ to the ground truth tokens. On the other hand, the sequence length of $Y_{CE}$ is determined by linguistic representation $H_L$, so we use CE criterion to align the text sequence of  $Y_{CE}$ to the ground truth transcript.

The different aggregation mechanisms including Gated Acoustic Guided Attention, Gated Linguistic-Guided Attention and Gated Cross-Modal Attention are evaluated and compared in Table~\ref{tab:ablation_fea_agg}.

\subsubsection{Embedding Attention Module}
\label{sec:embed_fusion}
Recent works~\cite{yi2021efciently, yu2021non} directly connect the BERT on the top of the acoustic encoder and simply replace the embedding layer with the acoustic features generated by the acoustic encoder, causing the catastrophic forgetting problem.

To fill the gap of embedding inconsistency, we propose the Embedding Attention Module and insert it behind the embedding layer of BERT to incorporate the acoustic information into the word embedding instead of simply replacing them. 
We first introduce a Gated Attention operation in this module.
As shown in Figure~\ref{fig:overview},  word embedding ${E}$ generated by embedding layer is fed to a self-attention layer followed by a feed-forward layer to capture  higher level linguistic embedding $E_L$. Then, a  multi-head self-attention followed by a gated weighting layer takes $E_L$ as the query vector and  acoustic embedding $H_A$  generated by wav2vec 2.0 as the key vector as well as value vector to fuse the linguistic embedding and acoustic embedding. Thus, as a conditional masked language model, BERT can learn to predict the masked word under the conditional acoustic information and provided enhanced linguistic representation. 

Furthermore, for the input of the embedding layer of BERT, it is easy to overfit when using ground truth transcripts while it is hard to converge when using transcripts predicted by wav2vec2.0 encoder. To solve this issue, we propose a "Sampling with Decay" mechanism by feeding BERT either the masked ground truth transcript $Y^r$ or the predicted CTC result $Y_{CTC_1}$ with a certain probability during training. The probability $p$ of selecting from $Y^r$ decreases linearly as the number of training steps increases.

Through the Embedding Attention Module with "Sampling with Decay" mechanism,  we further integrate the acoustic and linguistic information from the embedding level to facilitate better fusion between wav2vec 2.0 encoder and BERT encoder. Table~\ref{tab:ablation_embed_fusion} verifies the effectiveness of each component of our proposed Embedding Attention Module.

\subsubsection{Inference}
\label{sec:infer}

For inference, we first feed the result $Y_{CTC_1}$ into BERT encoder; then select the one with higher confidence from the two outputs $Y_{CTC_2}$ and $Y_{CE}$ as our final output. 
\section{Experiments}

\begin{table}[]
\setlength\tabcolsep{2pt}
\centering
\tiny
\caption{Results of low resource ASR on IARPA BABEL in terms of CER (\%).}
\begin{tabular}{@{}l|c|llll@{}}
\toprule
{Method} & {Pre-trained} & {Vi}     & {Sw}    &  {Ta}  &  {Avg}   \\ \hline
Mono-BLSTMP~\cite{Cho2018MultilingualSS} & \multirow{3}{*}{-} &54.3 & 33.1 & 55.3 & 47.6 \\
Multi-BLSTMP~\cite{Cho2018MultilingualSS} & {} &41.0 & - & 48.5  & 44.8 \\
Multi-BLSTMP+ VGG~\cite{Cho2018MultilingualSS} &{} & 37.4 & - & 45.5 & 41.5 \\ \hline \hline
wav2vec 2.0~\cite{baevski2020wav2vec} &\multirow{4}{*}{\tabincell{c}{wav2vec 2.0 \\ (Base)}} &  21.8 & 15.5 & 29.3 & 22.2 \\
wav2vec 2.0 w/ 4-gram~\cite{baevski2020wav2vec} &{} &  21.1 & 14.9 & 29.9 & 22.0 \\ 
XLSR-Monolingual~\cite{conneau2020unsupervised} &{} &  25.2 & 26.8 & 36.0 & 29.3 \\
XLSR-10~\cite{conneau2020unsupervised} &{} &  21.7 & 16.6 & 30.5 & 22.9 \\ \hline
BERT rescoring~\cite{shin2019effective} &\multirow{3}{*}{\tabincell{c}{w/ mBERT}} &  21.3 & 15.3 & 29.1  & 21.9 \\
Adapter-BERT~\cite{NEURIPS2020_7a6a74cb} &{} &  22.5 & 17.6 & 29.8 & 23.3 \\
w2v-cif-bert~\cite{yi2021efciently} &{} &  24.1 & 21.5 & 41.9 & 29.2 \\  
\textbf{our Wav-BERT} &{} &  \textbf{19.5} & \textbf{14.8} & \textbf{28.8}  & \textbf{21.0} \\ \hline  \hline
XLSR-10~\cite{conneau2020unsupervised} &\multirow{2}{*}{\tabincell{c}{wav2vec 2.0 \\ (Large) }} &  19.9 & 14.9 & 28.6 & 21.1 \\ 
XLSR-53~\cite{conneau2020unsupervised} &{} &  21.8 & 21.3 & \textbf{27.4} & 23.5 \\ \hline
\textbf{our Wav-BERT w/ XLSR-53} &{w/ mBERT} &  \textbf{19.3} & \textbf{13.8} & 28.0 &  \textbf{20.4} \\ \bottomrule
\end{tabular}
\label{tab:sota_babel}
\end{table}

In this section, we first illustrate the implementation details of our Wav-BERT. Then we introduce two low-resource speech recognition datasets containing several languages as well as the comparison results among our approach and baseline methods. Furthermore, we conduct ablation studies to validate the effectiveness of each main component of our Wav-BERT and present some case studies for perceptual comparison.

\noindent\textbf{Implementation Details.}
For our proposed Representation Aggregation Module and Embedding Attention Module, 
the heads and embedding dimensions of all multi-head attention are set to 8 and 768 respectively. 
Meanwhile, the inner-layer dimension of the position-wise feed-forward is set to 2048. Regarding optimization details, we train our model as well as baselines based on wav2vec 2.0 Base for 200K steps with one GeForce RTX 3090 GPU, setting max tokens and update frequency to 640000 and 4 correspondingly. As for experiments using XLSR-53~\cite{conneau2020unsupervised}, three GeForce RTX 3090 GPUs are used with max tokens as 480000 and update frequency as 4. We use the three-stage learning rate policy with the initial learning rate as 5e-5, and set each stage ratio to 0.05, 0.45 and 0.5. Besides, we set the weight $\mu_{1}$, $\mu_{2}$, $\mu_{3}$ and $\mu_{4}$ for each loss to 0.5 for training. Other optimizer settings are the same as wav2vec 2.0~\cite{baevski2020wav2vec}. In terms of the "Sampling with Decay" policy, languages in  IARPA BABEL start from 100K steps to 200K steps, while in AISHELL-1 it starts from 40k steps to 100k steps, all with $p$ decreasing from 90\% to 10\%. 

\noindent\textbf{Datasets.}
IARPA BABEL~\cite{gales2014speech} is an open-source multilingual corpus of conversational telephone speech. For low resource evaluation, we randomly select 3 kinds of languages with few data: Swahili (Sw), Tamil (Ta) and Vietnamese (Vi). We adopt the same setup as~\cite{conneau2020unsupervised} and use the dev folder of the BABEL dataset as our test set since "eval" data are not released. We re-sample audios of all languages to 16kHz. AISHELL-1~\cite{bu2017aishell} is an open-source and high-quality Mandarin speech corpus, and is widely used in the speech community, which contains 178 hours of Mandarin speech data. Although the data is in Chinese, a common used language, the quantity is small. Thus, it can also verify our Wav-BERT for low-resource data. Moreover, there are many latest state-of-the-art methods on this dataset to be compared. 

For a fair comparison, we use the official wav2vec 2.0 (Base/Large) model, XLSR-53, and mBERT models as the initial encoders. All model checkpoint download links are described in the  appendix.


\subsection{Results on IARPA BABEL}

\begin{table}[]
\setlength\tabcolsep{3pt} 
\centering
\scriptsize
\caption{Results of  ASR on AISHELL-1 in terms of CER(\%).}
\begin{tabular}{@{}l|c|ll@{}}
\toprule
\multirow{2}{*}{Method} & \multirow{2}{*}{Pre-trained}& \multicolumn{2}{c}{AISHELL-1}                                              \\ \cline{3-4}
                          &  & dev            & test                  \\ \hline
Kaldi chain~\cite{yu2021non} & \multirow{11}{*}{-} & - & 7.5 \\
Kaldi nnet3~\cite{yu2021non} & & - & 8.6 \\
LAS~\cite{shan2019component} & & - & 10.6 \\
ESPnet (Transformer)~\cite{karita2019comparative} & & 6.0 & 6.7 \\
SA-T~\cite{tian2019self} & & 8.3 & 9.3 \\
SAN-M~\cite{gao2020san} & & 5.7 & 6.5 \\
CAT~\cite{an2019cat} & & - & 6.3 \\
LFML~\cite{chen2019listen} & & 6.2 & 6.7 \\
LASO~\cite{bai2021fast} & & 5.9 & 6.9 \\
NAR-Transformer~\cite{song2020non} & & 5.6 & 6.3 \\
Wenet~\cite{zhang2020unified} &  & - & 4.7 \\ \hline \hline
LASO with BERT~\cite{bai2021fast} & \multirow{2}{*}{BERT} & 5.3 & 6.1 \\
NAR-BERT-ASR~\cite{yu2021non} & & 4.9 & 5.5 \\  \hline \hline
wav2vec 2.0~\cite{baevski2020wav2vec} &\multirow{3}{*} {wav2vec 2.0} & 7.9 & 8.4 \\ 
wav2vec 2.0 (cn)~\cite{baevski2020wav2vec} & & 5.2 & 5.8 \\ 
wav2vec 2.0 (cn) w/ 4-gram~\cite{baevski2020wav2vec} & & 4.5 & 4.9 \\ \hline \hline
BERT rescoring~\cite{shin2019effective} & & 4.2 & 4.5 \\ 
Adapter-BERT~\cite{NEURIPS2020_7a6a74cb} & wav2vec 2.0 & 6.9 & 7.3 \\ 
w2v-cif-bert~\cite{yi2021efciently} & w/ BERT & 5.6 & 6.3 \\
\textbf{our Wav-BERT w/ wav2vec 2.0} & & \textbf{3.8} & \textbf{4.0} \\ 
\textbf{our Wav-BERT w/ wav2vec 2.0 (cn)} & & \textbf{3.6} & \textbf{3.8}  \\  \bottomrule 
\end{tabular}
\label{tab:sota_aishell}
\end{table}

Table~\ref{tab:sota_babel} reports the results on IARPA BABEL in terms of character error rate (CER), where our Wav-BERT achieves state-of-the-art performance on all low-resource languages. We find some interesting points comparing the results. First, the performance of the methods without pre-training is quite bad, which indicates that the conventional end-to-end models are impractical for low-resource languages due to the limited data. Second, the pre-training models like wav2vec 2.0 and XLSR largely improve the recognition accuracy thanks to the powerful acoustic representation learned from the huge amount of high-resource language data. Third, in addition to the pre-trained acoustic model, other methods also utilize a pre-trained language model like mBERT while the results change slightly or even become worse. One of the reasons is that the methods that construct adapters in BERT (ADapter-BERT) or simply combine BERT with wav2vec 2.0 (w2v-cif-bert) inevitably suffer from the embedding inconsistency problem and fail to make the best use of pre-trained linguistic representation. As for our Wav-BERT, it effectively facilitates the cooperation of the pre-trained acoustic and language models by the proposed fusion modules from representation level to embedding level. As a result, it can consistently improve the ASR results for different low-resource languages. Moreover, when the pre-trained model (e.g. wav2vec 2.0) becomes larger, the performance of our Wav-BERT will be also improved while it requires more GPU resources to tune the whole model.  



\subsection{Results on AISHELL-1}

\begin{table}[]
\setlength\tabcolsep{3.5pt}
\centering
\scriptsize
\caption{Results of different components in Representation Aggregation Module for ASR on IARPA BABEL and AISHELL-1 named CN in terms of CER(\%).}
\begin{tabular}{@{}l|llccl@{}}
\toprule
{Method} & {Vi}          & {Sw}      & {CN-dev} & {CN-test} & {Avg}           \\  \hline
{Gated Cross-Modal Attention} &  \textbf{19.5}  & \textbf{14.8} & \textbf{3.8} &  \textbf{4.0} & \textbf{10.5} \\ 
w/o Gated Weighting    & 	    19.6            & 	 14.9        & 3.9 &  4.2  &  10.7      \\ \hline
Gated Acoustic-Guided Attention  &       20.4    	          &    15.0       & 4.4   & 4.7 &  11.1   \\
Gated Linguistic-Guided Attention    &  25.6             &     18.3      & 5.7 &  6.4 & 14.0 \\\bottomrule
\end{tabular}
\label{tab:ablation_fea_agg}
\end{table}

\begin{table}[]
\centering
\scriptsize

\caption{Results of different components in  Embedding Attention Module for ASR on IARPA BABEL and AISHELL-1 named CN in terms of CER(\%).}
\begin{tabular}{@{}l|llccl@{}}
\toprule
{Method} & {Vi}          & {Sw}      & {CN-dev} & {CN-test}    & {Avg}                 \\ \hline
Embedding Replacement  &    21.1      & 	15.4          &      6.0    &     6.4  & 12.2   \\ \hline
{our Embedding Attention} & \textbf{19.5} & \textbf{14.8} & \textbf{3.8} & \textbf{4.0} & \textbf{10.5} \\
w/o Sampling with Decay    & 	 22.0      & 	  15.7      & 	5.7          &   6.2  & 12.4       \\
w/o Gated Attention    &      20.7   &    15.3      &      4.1  &   4.3  & 11.1   \\ 
\bottomrule
\end{tabular}
\label{tab:ablation_embed_fusion}
\end{table}

\begin{table*}[ht]
\setlength\tabcolsep{3pt} 
\centering
\tiny
\caption{Predicted examples on AISHELL-1 test set generated by Wav2vec 2.0, BERT rescoring, w2v-cif-bert and our Wav-BERT. The differences words are marked with pronunciation. The wrong words are marked in red. The translations of the sentences are also provided.}
\begin{tabular}{@{}l|llll@{}}
\toprule
{Method}    & {Predicted example with translation}     \\ \hline 
\raisebox{-.5\height}{wav2vec 2.0~\cite{baevski2020wav2vec}}   & \raisebox{-.5\height}{\includegraphics[scale=2]{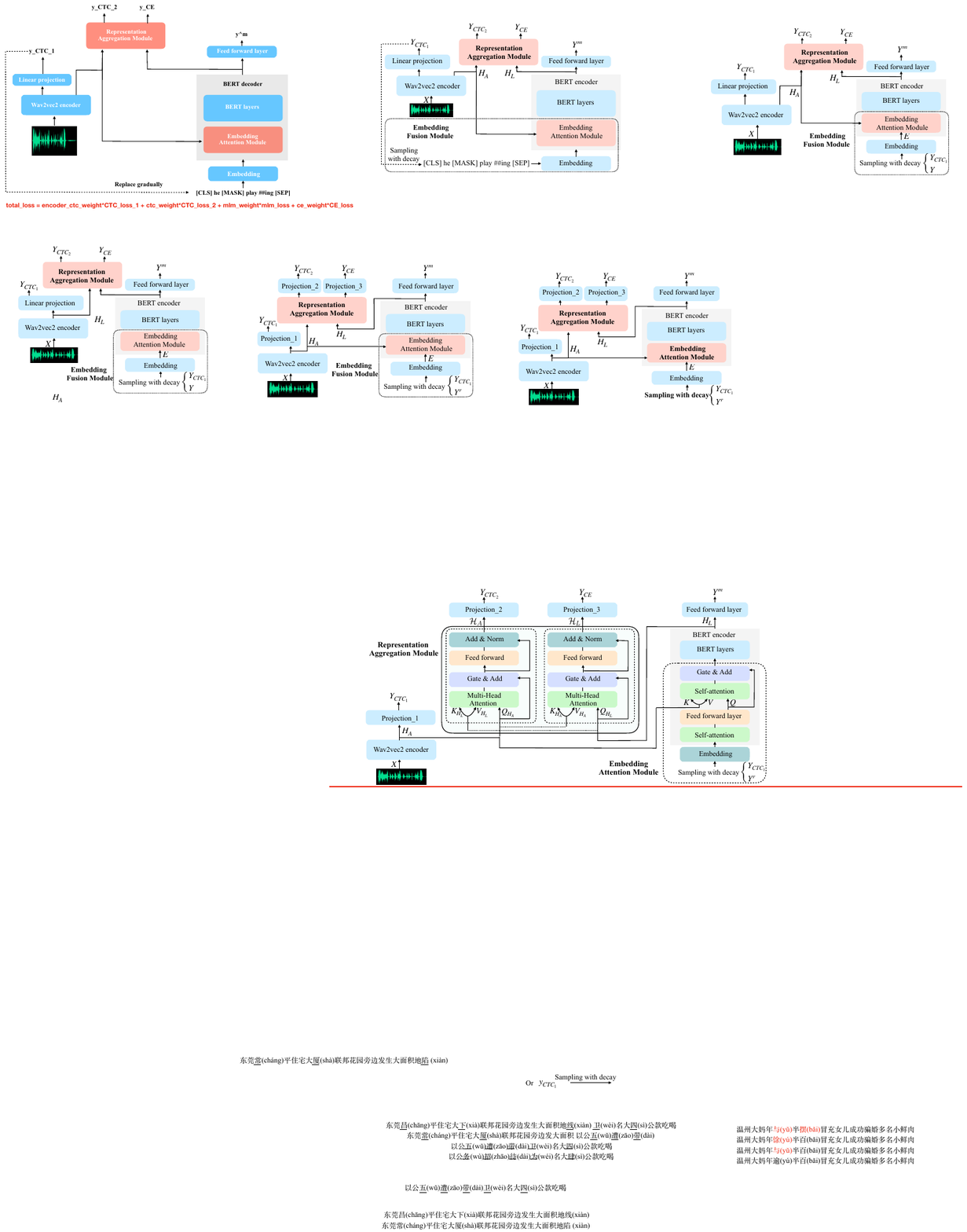}} \\ 
& Wenzhou aunt Nian and banpai pretended to be their daughter and got married successfully. \\ \hline
\raisebox{-.5\height}{BERT rescoring~\cite{shin2019effective}}  & \raisebox{-.5\height}{\includegraphics[scale=2]{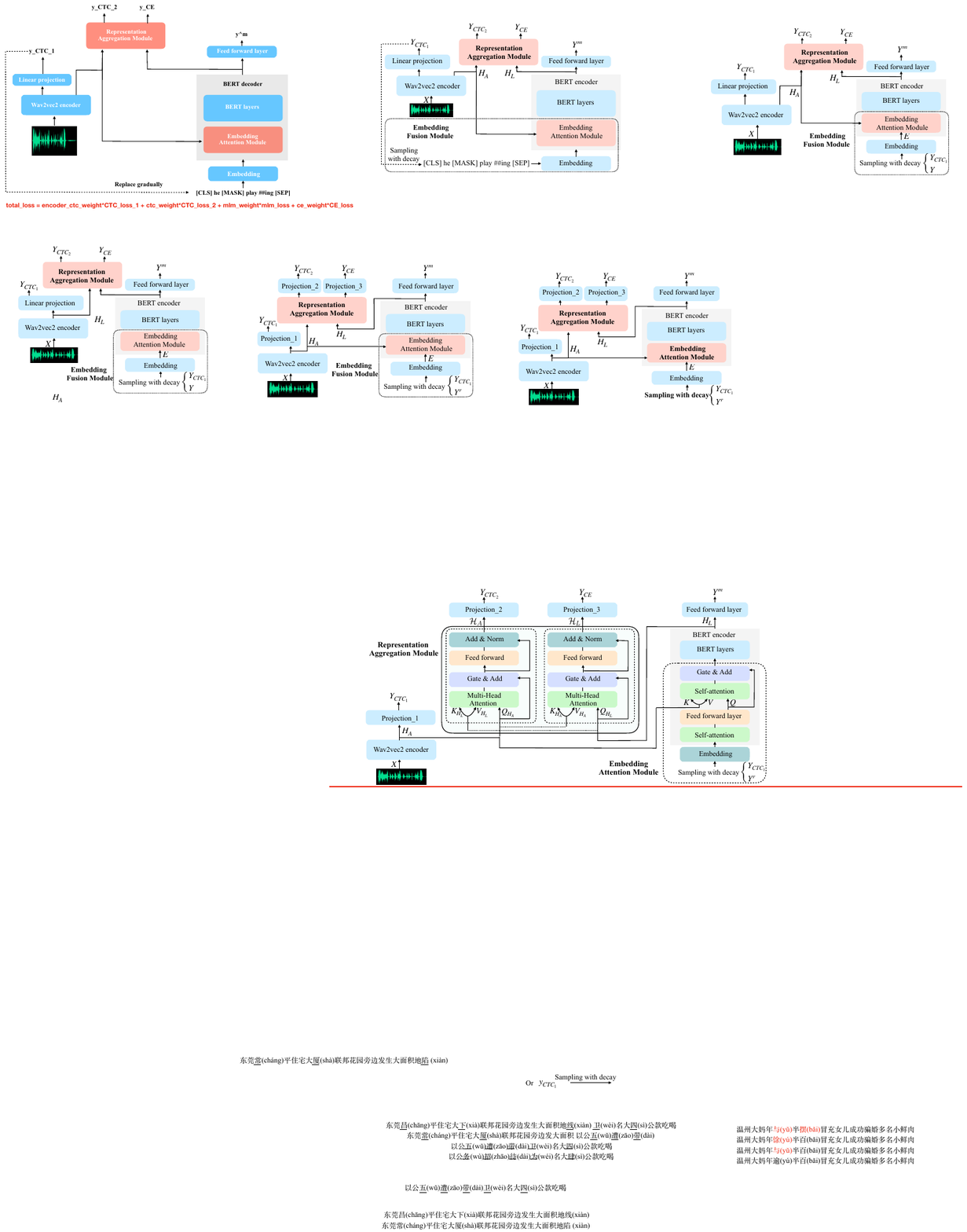}} \\ 
& More than half of Wenzhou's old aunt pretended to be her daughter and successfully cheated many young people into marriage.. \\ \hline
\raisebox{-.5\height}{w2v-cif-bert~\cite{yi2021efciently}}  & \raisebox{-.5\height}{\includegraphics[scale=2]{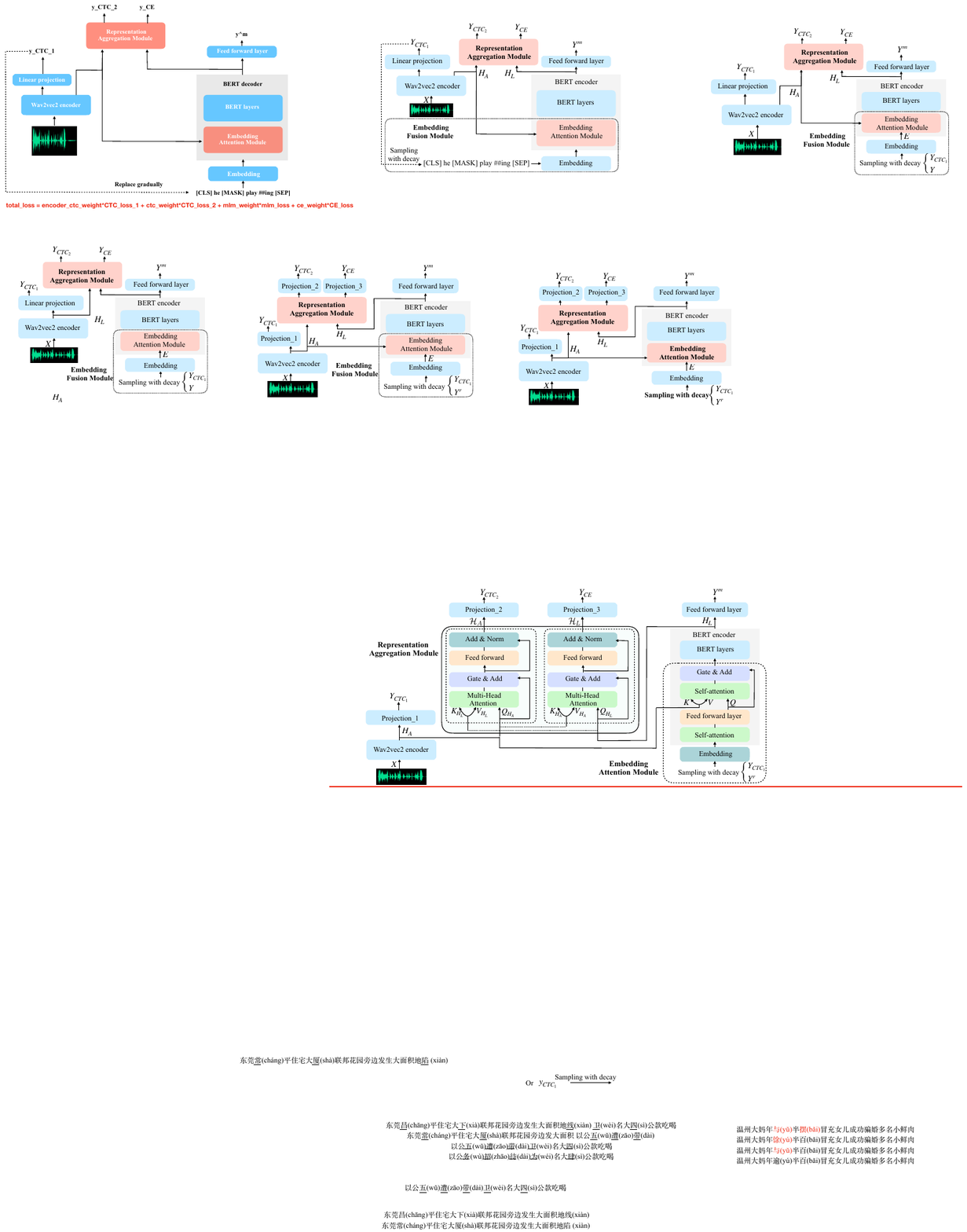}} \\ 
& Wenzhou aunt year and half a hundred pretending to be daughters have successfully cheated into marriage, and there are many young people. \\ \hline
\raisebox{-.5\height}{\textbf{our Wav-BERT}}  & \raisebox{-.5\height}{\includegraphics[scale=2]{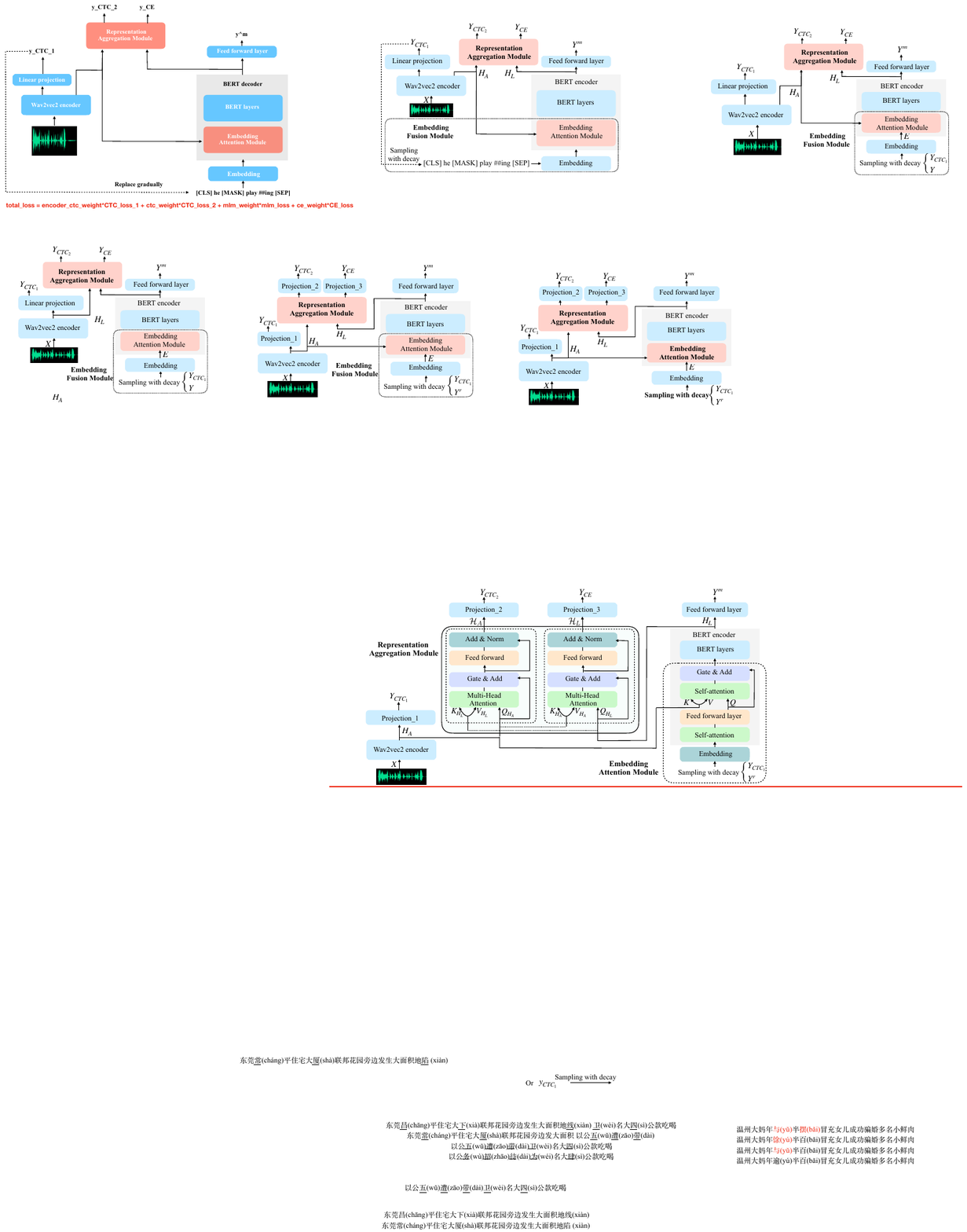}}	\\ 
& Wenzhou aunt is more than half a hundred years old, pretending to be her daughter, and has successfully cheated many young people into marriage. \\
\bottomrule
\end{tabular}
\label{tab:case_study}
\end{table*}

Table~\ref{tab:sota_aishell} reports the comparison results on AISHELL-1. In addition to the baselines mentioned above, we also report more latest works for comparison. The data quantity of this dataset is larger than that of IARPA BABEL, so all the methods perform much better. It also accounts for that the performance distance between the methods with pre-trained models and those without pre-trained models becomes small. During the methods without pre-trained models, wenet~\cite{zhang2020unified} achieves the best results due to its advanced CTC-Conformer~\cite{graves2006connectionist,gulati2020conformer} architecture, better attention rescoring decoding strategy and larger training epoch number. With the pre-trained language  model of BERT, NAR-BERT-ASR~\cite{yu2021non} stacked a decoder initialized by a pre-trained BERT model on the top of the transformer encoder and achieves competitive results on AISHELL-1. Regarding methods using the pre-trained acoustic model, the official wav2vec 2.0 Base model that pre-trained on 960 hours of Librispeech corpus achieves great results as the model learned good representations of speech. Furthermore, we also collect and use 1960 hours of public Mandarin speech data to pre-train a wav2vec 2.0 (cn) model, which obtains better performance on AISHELL-1 evaluation.  In conclusion, our Wav-BERT not only improves the performance of both wav2vec 2.0 and wav2vec 2.0 (cn) models, but also outperforms other state-of-the-art methods unifying wav2vec 2.0 and BERT. It further demonstrates the generalization of Wav-BERT on different low-resource ASR datasets with different data sizes.

\subsection{Comparison of model fusion methods}

As illustrate in Section~\ref{sec:model_fusion}, there are many different model fusion methods to fuse the pre-trained wav2vec 2.0 and BERT. We compare our Wav-BERT with these methods and report the results in Table~\ref{tab:sota_babel} and Table~\ref{tab:sota_aishell}. First, by using BERT to rescore $N$-best hypotheses generated by wav2vec 2.0 with CTC beam search, rescoring~\cite{shin2019effective} (Figure~\ref{fig:model_fusion} (a)) is slightly better than wav2vec 2.0, but its inference process is time-consuming. Second, w2v-cif-bert~\cite{yi2021efciently} uses CIF to connect wav2vec 2.0 and BERT in a cascade way and replace word embedding with acoustic embedding as input for BERT. It is better than wav2vec 2.0 in AISHELL-1 but worse in BABEL for the reason that the mBERT is not as well trained as the bert-base-chinese model, resulting in a more severe catastrophic forgetting problem after replacing its input. Third, Adapter-BERT that inserts adapter modules into each BERT layer and tunes it on the training data, has an inconspicuous improvement or even performance degradation since the insertion of adapters affects the pre-trained representation of BERT. Finally, our Wav-BERT signiﬁcantly surpasses other methods, which indicates that our model can effectively exploit the acoustic and linguistic information through the multi-level hierarchical fusion. Besides, our cooperative learning methods can also help the pre-trained encoders to avoid catastrophic forgetting of pre-training information so that the whole model can converge faster and better.

\subsection{Ablation Studies}

\subsubsection{Representation Aggregation Module}

To investigate the effectiveness of our Representation Aggregation Module, we present results for Gated Linguistic-Guided Attention, Gated Acoustic-Guided Attention, removing gated weighting in Table~\ref{tab:ablation_fea_agg}. We can find that the effect of gated weighting, while small, is still existent, which can automatically measure the importance of the acoustic and linguistic representation while aggregating those two kinds of representation. 
Compared with Gated Cross-Modal Attention, Gated Acoustic-Guided Attention and Gated Linguistic-Guided Attention increases the average CER by 0.6\% and 3.5\% respectively, which indicates that the attention in each direction plays an important role in our Representation Aggregation Module while Gated Acoustic-Guided Attention makes a greater contribution since speech recognition task is more dependent on acoustic information.
\subsubsection{Embedding Attention Module}
The results in Table~\ref{tab:ablation_embed_fusion} further verify the effectiveness of our Embedding Attention Module. First, we report the result of Embedding Replacement that simply replaces the original word embedding with the acoustic embedding as the input of BERT like previous works~\cite{yu2021non}. As expected, the performance is poor especially on AISHELL-1, which indicates that such simple replacement methods will be affected by the embedding inconsistency problem. In contrast, we solve this challenge by the proposed Embedding Attention Module including the sampling mechanism and Gated Attention, so that the performance is largely improved. Second, when turning off "Sampling with Decay" or Gated Attention, the average CER increased by 1.9\% and 0.6\% respectively.
It demonstrates that the "Sampling with Decay" mechanism effectively alleviates the embedding inconsistency of BERT between inference and training. Mover, the Gated Attention effectively provides additional acoustic information to the input of BERT, facilitating it to capture more reliable linguistic representation.



\subsection{Case Studies}
We further present some case studies in Table~\ref{tab:case_study}, to illustrate the importance of acoustic and linguistic information for speech recognition. We provided some transcript examples obtained from the baseline methods and our Wav-BERT with the same input from AISHELL-1 test set. The pronunciations of the keywords and the English translation of the whole sentence are also provided. 
As can be observed, all the baseline methods predict one or two wrong words with similar pronunciation as the wrong words, which leads to an unreasonable sentence. On the contrary, thanks to the cooperative learning of acoustic and linguistic information, our Wav-BERT can successfully recognize the whole sentence without any word error.


\section{Conclusion}
In this work, based on the powerful wav2vec 2.0 and BERT models, we introduce cooperative acoustic and linguistic representation learning for low-resource speech recognition. To solve the representation discrepancy and embedding inconsistency challenges, we design a Representation Aggregation Module and an Embedding Attention Module to facilitate the cooperation of the two pre-trained models and thus boost the representation learning. Extensive experimental results demonstrate that our proposed Wav-BERT can significantly improve low-resource ASR performances in different languages.  In future work, we
will investigate more effective modules to infuse more types of knowledge, and apply our framework to more pre-trained models to promote the development of low-resource speech tasks.

\section*{Acknowledgement}
This work was supported in part by National Key R\&D Program of China under Grant No. 2020AAA0109700, National Natural Science Foundation of China (NSFC) under Grant No.U19A2073 and No.61976233, Guangdong Province Basic and Applied Basic Research (Regional Joint Fund-Key) Grant No.2019B1515120039, Guangdong Outstanding Youth Fund (Grant No. 2021B1515020061), Shenzhen Fundamental Research Program (Project No. RCYX20200714114642083, No. JCYJ20190807154211365).

\bibliography{anthology,custom}
\bibliographystyle{acl_natbib}

\appendix
\section{Datasets}
Both IARPA BABEL dataset~\cite{gales2014speech} and AISHELL-1~\cite{bu2017aishell} are open-source and high-quality speech datasets, and are widely used in the speech community. Among them, AISHELL-1 can be downloaded for free here\footnote{\url{https://www.openslr.org/33/}}, For each speaker in it, around 360 utterances(about 26 minutes of speech) are released. Table~\ref{tab:aishell} provides a summary
of all subsets in the corpus. As for IARPA BABEL, it can be purchased through LDC\footnote{\url{https://www.ldc.upenn.edu/}}(eg. Vietnamese Language Pack\footnote{\url{https://catalog.ldc.upenn.edu/LDC2017S01}}). Table~\ref{tab:babel} summarizes the amount of data in hours for the language used in our experiments on the "Full Language Pack" (FLP) condition. Researchers can easily reproduce or compare our results with the same languages.

\begin{table}[h]
    \centering
    \caption{AISHELL-1 dataset statistics.}
    \begin{tabular}{l|l|l|l}
    \toprule
    Subset      & Duration(hrs) & Male & Female  \\ \hline
    Training    & 150           & 161 & 179 \\ \hline 
    Development & 10            & 12  & 28\\  \hline 
    Test        & 5             & 13  & 7 \\  \hline 
    \end{tabular}
    \label{tab:aishell}
\end{table}

\begin{table}[h]
    \centering
    \caption{IARPA BABEL dataset statistics.}
    \begin{tabular}{l|l|l}
    \toprule
    Language   & Train(hrs) & Eval(hrs)  \\ \hline
    Vietnamese & 87.72 & 11.00 \\ \cline{1-3}  
    Swahili    & 44.39 & 10.65 \\ \cline{1-3}  
    Tamil      & 69.35 & 11.68 \\ \bottomrule
    \end{tabular}
    \label{tab:babel}
\end{table}

\section{Ours Wav-BERT Model}
Our model checkpoint described in Sec 5 can be downloaded \href{https://drive.google.com/drive/folders/1lISXln7nzeOiuWPuyHOo_jBgtWKH77jS?usp=sharing}{here}. With limited storage space, thus we only upload the model using wav2vec 2.0 Base. 

\section{Pre-trained Models}
We use different pre-trained acoutic and language models in our experiment described in Sec 5. All of them are open-source except the wav2vec 2.0~\cite{baevski2020wav2vec} pre-trained in Chinese by ourselves. For pre-trained language models, the bert-base-chinese model can be download here~\footnote{\url{https://s3.amazonaws.com/models.huggingface.co/bert/bert-base-chinese.tar.gz}}, and the multilingual mBERT can be download here~\footnote{\url{https://s3.amazonaws.com/models.huggingface.co/bert/bert-base-multilingual-uncased.tar.gz}}. For pre-trained acoustic models, the official wav2vec 2.0 pre-trained on English can be download here~\footnote{\url{https://dl.fbaipublicfiles.com/fairseq/wav2vec/wav2vec_small.pt}}, and the XLSR-53~\cite{conneau2020unsupervised} model can be downloaded here~\footnote{\url{https://dl.fbaipublicfiles.com/fairseq/wav2vec/xlsr_53_56k.pt}}. Besides, though the wav2vec 2.0(cn) pre-trained on 1,960 hours of Chinese data cannot open-source, both the used training code and datasets are open-source, which means researchers still can reproduce our results. In details, we base on the Fairseq framework~\footnote{\url{https://github.com/pytorch/fairseq}}~\cite{ott2019fairseq} to pre-train our model 8 GeForce RTX 3090 GPUs with max tokens and update frequency setting to 1400000 and 8 respectively, consuming about one week to train 400K steps. Besides, the used datasets are DiDiSpeech~\cite{guo2021didispeech}, PVTC~\footnote{\url{https://www.pvtc2020.org/index.html}}, ST-CMDS~\footnote{\url{http://www.openslr.org/38/}}, aidatatang~\footnote{\url{http://www.openslr.org/62/}}, AISHELL-1, AISHELL-3~\cite{shi2020aishell}, MAGICDATA~\footnote{\url{http://www.openslr.org/68/}}, MagicDataSpeech~\footnote{\url{https://www.biendata.xyz/competition/magicdata/}}, Primewords~\footnote{\url{http://www.openslr.org/47/}} and Thchs~\footnote{~\url{http://www.openslr.org/18/}}.

\section{Baselines}
We describe some baseline methods below,  which are reproduced by ourselves or experimented with the open-source code.
\begin{enumerate}
    \item Wav2vec 2.0 w/ 4-gram: For each language, results from the trained wav2vec 2.0 model with beam search, are rescored by the 4-gram language model. Specifically, the 4-gram model is trained by transcripts in the training set of each language, using the KenLM~\cite{heafield2011kenlm} framework. And the beam size for beam search is set to 50.
    \item BERT rescoring~\cite{chiu2021innovative,shin2019effective}: For each language, results from the trained wav2vec 2.0 model with beam search, are rescored by the fine-tuned language model(mBERT or bert-base-chinese model). Specifically, the linguistic decoder is fine-tuned by transcripts in the training set of each language using masked language model(MLM) objective~\cite{devlin2018bert} of BERT. In rescoring stage, we mask each word in the sentence once at a time, then sum all the log-likelihoods of the masked words from each masked input instance. Finally rescoring the sentence with both the likelihoods from acoustic and language model. Besides, considering it is time-consuming, the beam size for beam search is set to 5.
    \item Adapter-BERT: This method is inspired by AB-Net~\cite{NEURIPS2020_7a6a74cb}, cross-attention adapters are inserted to each BERT layer to unify the wav2vec 2.0 and BERT model. Output from the feed-forward layer at the last of BERT is supervised by the cross-entropy criterion. In inference, the Mask-Predict algorithm~\cite{ghazvininejad2019mask} is adopted.
    \item Embedding Replacement: Inspired by previous work~\cite{yu2021non}, we use similar architecture as it but replace the acoustic encoder with wav2vec 2.0 and keep our Representation Aggregation Module. We use position embeddings as query vector and acoustic representation from wav2vec 2.0 as key and value vector to attention block followed by 3 self-attention block, which is the same as ~\cite{yu2021non}, generating aligned acoustic representation $H_{pos}$. Then $H_{pos}$ is used as the input of BERT, replacing the word embedding. Finally, Representation Aggregation Module takes both the $H_{pos}$ and linguistic representation from BERT as input, just the same as our Wav-BERT. It is worth mention that the length of the position embedding is set to 60, considering it cost too much GPU memory for a larger value.
\end{enumerate}

\section{More Implementation Details}
Most of the significant experiment details are described in Sec 5. Aiming to let researcher reproduce our result more easily, we describe more details below. About the data augmentation, mask probability and mask channel probability are set to 0.65 and 0.5 respectively the same as setting in wav2vec 2.0~\cite{baevski2020wav2vec}  for 100 hour training data. Besides, we use adam optimizer, setting adam betas and adam eps to (0.9,0.98) and 1e-08 individually. In data preprocessing, we use feature normalize for wav2vec 2.0 Base model but not for the XLSR-53 model, keeping consistent with the pre-training setting. Also, we filter some samples whose length of speech shorter than 0.5 seconds as well as number of subwords less than 1 or bigger than 512 in training set. Regarding the training time, training our Wav-BERT model with wav2vec 2.0 Base model spends less than 2 days, and 5 days with the XLSR-53 model. Finally, the number of parameters in our model with wav2vec 2.0 Base is about 380M, and 600M with XLSR-53, which is slightly different with different languages.





\end{document}